\documentclass[lettersize,journal]{IEEEtran}
\usepackage{amsmath,amsfonts}
\usepackage{algorithmic}
\usepackage{algorithm}
\usepackage{array}
\usepackage[caption=false,font=normalsize,labelfont=sf,textfont=sf]{subfig}
\usepackage{textcomp}
\usepackage{stfloats}
\usepackage{url}
\usepackage{verbatim}
\usepackage{graphicx}
\usepackage{cite}
\usepackage{algorithm}
\usepackage{algorithmic}

\usepackage{comment}
\usepackage{amsmath,amssymb}
\usepackage{color}
\usepackage{multirow}
\usepackage{booktabs}
\usepackage{amssymb}
\usepackage{color}

\begin{document}

\title{Scene Text Image Super-Resolution via Content Perceptual Loss and Criss-Cross Transformer Blocks}

\author{
Rui Qin\hspace{0.5in}
Bin Wang\hspace{0.5in}
Yu-Wing Tai
}

\maketitle

\begin{abstract}
Text image super-resolution is a unique and important task to enhance readability of text images to humans. It is widely used as pre-processing in scene text recognition. However, due to the complex degradation in natural scenes, recovering high-resolution texts from the low-resolution inputs is ambiguous and challenging. Existing methods mainly leverage deep neural networks trained with pixel-wise losses designed for natural image reconstruction, which ignore the unique character characteristics of texts. A few works proposed content-based losses. However, they only focus on text recognizers' accuracy, while the reconstructed images may still be ambiguous to humans. Further, they often have weak generalizability to handle cross languages. 
To this end, we present TATSR, a Text-Aware Text Super-Resolution framework, which effectively learns the unique text characteristics using Criss-Cross Transformer Blocks (CCTBs) and a novel Content Perceptual (CP) Loss. 
The CCTB extracts vertical and horizontal content information from text images by two orthogonal transformers, respectively.
The CP Loss supervises the text reconstruction with content semantics by multi-scale text recognition features, which effectively incorporates content awareness into the framework. Extensive experiments on various language datasets demonstrate that TATSR outperforms state-of-the-art methods in terms of both recognition accuracy and human perception.
\end{abstract}

\begin{IEEEkeywords}
Super Rsolution, Scene Text Recognition (STR), Text Image, Attention, Convolutional Neural network
\end{IEEEkeywords}

\section{Introduction\label{sec:intro}}
Scene Text Recognition (STR) aims to extract character sequences from real-world scene text images. It has attracted much attention due to its importance in various scene-based text-related tasks, such as document retrieval~\cite{7781677} and license plate recognition~\cite{licenseDec}. However, in real-world scenarios, texts often occupy a small region and can be blurry, which leads to poor STR performance. To tackle these challenges, researchers investigate the Scene Text Image Super-Resolution (STISR) to enhance the quality of low-resolution text images for better recognition performance.
Previous works show that the widely-used pixel-wise losses in natural image super-resolution are not suitable for STISR because they are insensitive to text contexts (character shape and text sequence information).
To better match the text data, a few recent studies bring text-aware characteristics into supervision for better restoration and recognition performance. Scene Text Telescope (STT,~\cite{scenetexttelescope}) and the follow-up works~\cite{TATT,TG} perform text recognition on the super-resolution results and supervise the outputs and attention maps using the ground truth text labels.

\begin{figure}[h]
	\includegraphics[width=0.49\textwidth]{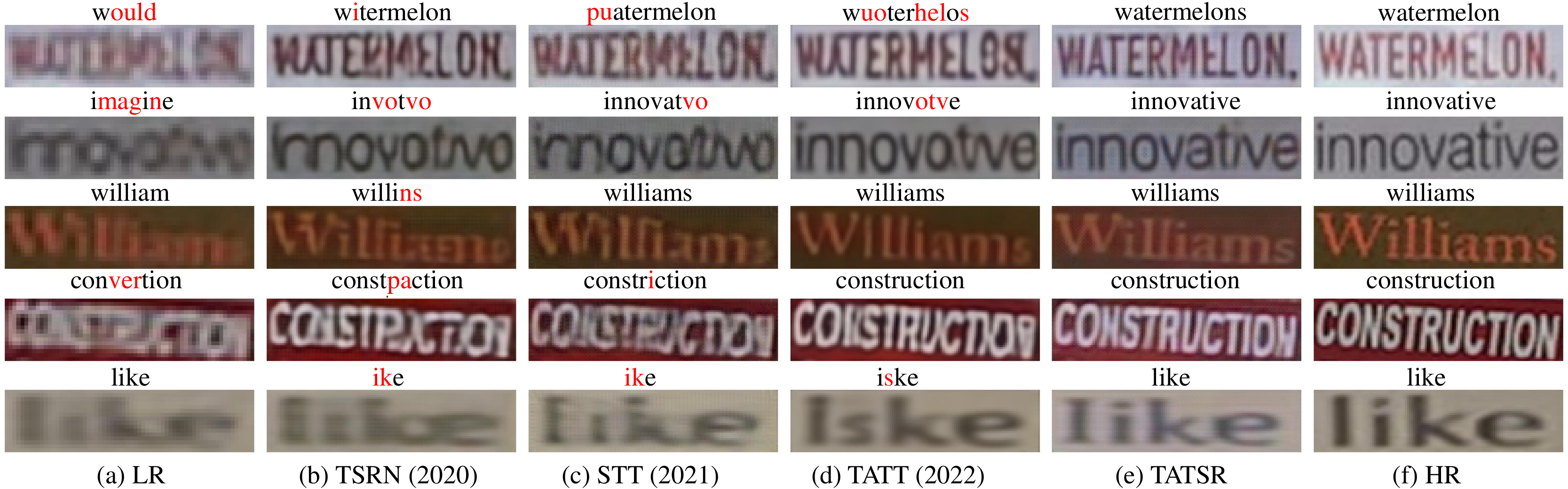}
	\caption{STISR aims to estimate high resolution text images from its degraded low resolution inputs. Our proposed TATSR can estimate the most recognizable characters with the clearest text strokes compared with other state-of-the-art STISR methods. The recognized text labels are shown above each image.} 
	\label{fig:first_page_cmp}

\end{figure}

However, we find that \textit{1)} the supervision using text recognition accuracy makes the trained network tend to maximize the character \textit{probabilities}, which is in conflict with the pixel-wise losses that maximize the image \textit{similarity} with the ground truth text images. 
\textit{2)} The recognizer’s feedback is not always accurate.
The inconsistency between the two kinds of losses makes the training unstable, leading to suboptimal results.
\textit{3)} The recognition-result-based losses are strongly limited by the language type, leading to inapplicability across different languages.
\textit{4)} The supervision with only the recognizer's feedback is insensitive to the local structure of characters (\textit{e.g.,} shapes and strokes), resulting in poor visual quality (\textit{e.g.,} Fig.~\ref{fig:first_page_cmp}). 

In order to tackle the above challenges, we propose a novel character-shape-sensitive text-aware loss, which is consistent with pixel-wise losses and cross-language utilizable, called Content Perceptual (CP) Loss. Specifically, we choose the recognizer's multi-scale features as the carrier of text information instead of the recognizer's final feedback to supervise the similarity between super-resolution results and high-resolution images in the text-oriented feature spaces. CP Loss can help the network to learn both precise character structure (from the local features) and text-aware content (from the global features) while simultaneously keeping the convergence point consistent with pixel-wise losses. Since parts of the text priors have cross-language similarities, such as the stroke layouts and the character shapes, our CP Loss can provide effective text-related supervision even in a cross-language scenario.


Besides text-oriented supervision, recent STISR methods use specific sequence processing modules to learn text priors, including character shape and text sequence information, mainly by Recurrent Neural Network (RNN) based modules. However, because RNN has limited memory, the RNN-based frameworks usually have performance drops when processing long text samples. To address this problem, we propose a new arbitrary-text-length-friendly sequence processing block called Criss-Cross Transformer Block (CCTB). Thanks to the global visibility of multi-head attention, the transformer-based CCTB can model content information of arbitrary-length texts and is especially beneficial in long text cases. Considering the common layout of text characters, we explicitly separate the learning of character shape and sequence information by the criss-cross sparse strategy, which simplifies the learning difficulty and also more text-content-appropriate than the rough dense transformer.

The contributions of our work can be summarized as follows:
\begin{enumerate}
	\item{We propose a novel character-shape-sensitive text-aware loss, which is consistent with pixel-wise losses and cross-language utilizable, called Content Perceptual (CP) Loss. It measures the distance between super-resolution results and ground truth in the multi-scale STR feature spaces, solving the problems in previous text-oriented losses.}
    \item{We propose a new sequence processing block called Criss-Cross Transformer Block (CCTB), which utilizes the transformer to model arbitrary-length text sequence information and applies a criss-cross sparse design to better appropriate the text data, addressing the RNN-based block's shortcoming on long text samples.}
	\item We construct a new STISR framework combining the CP Loss and CCTB, called Text-Aware Text Super-Resolution (TATSR). Experiments on various language datasets show that our proposed TATSR can effectively restore low-resolution scene text images and has visible improvements in visual perception and text recognition accuracy, achieving new state-of-the-art performance.
\end{enumerate}

\section{Related Work}
\subsection{Scene Text Recognition}

Traditional Scene Text Recognition (STR) methods often adopt a bottom-up strategy by recognizing each character first and then concatenating the whole word~\cite{Jaderberg2014deepfeature, he2016reading, su2014accurate}, or treating the task as a multi-classification task~\cite{jaderberg2016reading}. 

In contrast, recent solutions often adopt a top-down strategy that treats the scene text recognition task as an Image-to-Sequence task. Such a strategy effectively reduces the complexity of the model and makes it easier to treat text samples with variable lengths. According to the strategy of loss calculation, existing top-down methods can be divided into two categories, including the CTC-based methods and attention-based methods. CRNN~\cite{crnn} first proposed the RNN-based end-to-end scene text recognition method, which combined the Convolutional Neural Network (CNN) with Bi-directional Long Short-Term Memory (BLSTM, \cite{blstm}) for text feature extraction and introduced the Connectionist Temporal Classification (CTC,~\cite{graves2006connectionist}) Loss for alignment between feature sequences and text labels, making the sequence prediction unlimited by the fixed length of texts. Attention-based frameworks~\cite{shi2018aster,luo2019moran,cheng2017focusing, liu2018char, liao2019scene, wang2020exploring}, represented by ASTER~\cite{shi2018aster}, have been recently studied because of their robustness to text data with diverse text lengths and shapes. ASTER~\cite{shi2018aster} utilizes Spatial Transformer Network (STN)~\cite{stn} for text image rectification and conducts recognition using an attention-based sequence-to-sequence model.

In practice, CNN features were widely used as the basic feature extractor for both top-down and bottom-up strategies. Thus, it is natural to consider the CNN features extracted from STR models have captured the information closely related to character shape and text sequences. Based on this inspiration, we build our Content Perceptual Loss.

\subsection{Scene Text Image Super Resolution}
Unlike the Single Image Super-Resolution (SISR)~\cite{chen2019joint, dai2019second, dong2015image, guo2020dual, kim2016accurate, lai2017deep, ledig2017photo, lim2017enhanced, niu2020single, wang2018esrgan, zhang2018image, wang2021real}, Scene Text Image Super-Resolution (STISR) focuses more on optimizing human eye perception and recognizability of text images. Except for the earlier work~\cite{mancas2007introduction, wang2019text, Dong2015BoostingOC,zhang2017cnn} migrated from SISR, recent STISR works, started by~\cite{wang2020scene}, focus on mining the text-specific contextual information such as character shape and text sequence.~\cite{wang2020scene} constructed the real text super-resolution dataset and proposed a reasonable evaluation metric (the accuracy of mainstream recognition methods on super-resolution images). The main focus points of recent STISR works include super-resolution models specific to text data and task-specific text-oriented losses.
\subsubsection{Text-Specific Super-Resolution Model}
Like Scene Text Recognition, recent STISR models~\cite{wang2019textsr,wang2020scene, scenetexttelescope,2021Parallelly,TG,TATT,C3-STISR} mainly combine CNN and sequence modeling modules together for end-to-end framework construction. CNN is responsible for image feature extraction, and the sequence modeling modules focus on the learning of text sequence contents. TSRN~\cite{wang2020scene} and its follow-up works~\cite{TG, TATT, C3-STISR} use BLSTM~\cite{graves2012long, blstm, gru} as the main sequence modeling module. However, due to the limitation of the hidden state structure, BLSTM can not handle samples with long texts well. PCAN~\cite{2021Parallelly} improve the network block structures attending channels. STT~\cite{scenetexttelescope} introduces the Transformer~\cite{vaswani2017attention} to replace the BLSTM-based sequence processing blocks. Due to its rough full-image attention calculation, the single dense transformer has to learn all text contexts (the character shape and the sequence information) from a totally-flattened pixel sequence. This increases learning difficulty resulting in only slight improvement compared to the BLSTM-based modules but with much larger computations.

\subsubsection{Text-Oriented Loss}
The previous text-oriented losses use the deep feedback of the text recognizer as a provider of text semantics and design loss calculation based on it. According to the type of the feedback used, they can be divided into deep-feature-based and recognition-result-based. The deep-feature-based losses use the recognizer's deep features for loss calculation. For instance, the Position-Aware (PA) Loss in STT~\cite{scenetexttelescope} calculates the L2 distance of the transformer's attention map. The recognition-result-based losses~\cite{scenetexttelescope,wang2019textsr, TG, C3-STISR, TATT} mainly evaluate the difference of distribution between recognition results of super-resolution images and ground truth text labels. 
As discussed in Sec.~\ref{sec:intro}, both types of losses have weak supervision on the fine structure and edges of strokes. Besides, since the recognition results are highly correlated with language types and recognizer's accuracy, recognition-result-based losses are naturally unusable across languages and have unavoidable convergence conflicts with pixel-wise losses.

\section{Method}
In this section, we introduce the proposed Text-Aware Text Super-Resolution (TATSR) framework. We first start with a brief overview of our framework. Then, we describe our Content Perceptual (CP) Loss and Criss-Cross Transformer Block (CCTB) in detail, respectively.

\begin{figure*}[t]
	\centering
	\includegraphics[width=\textwidth]{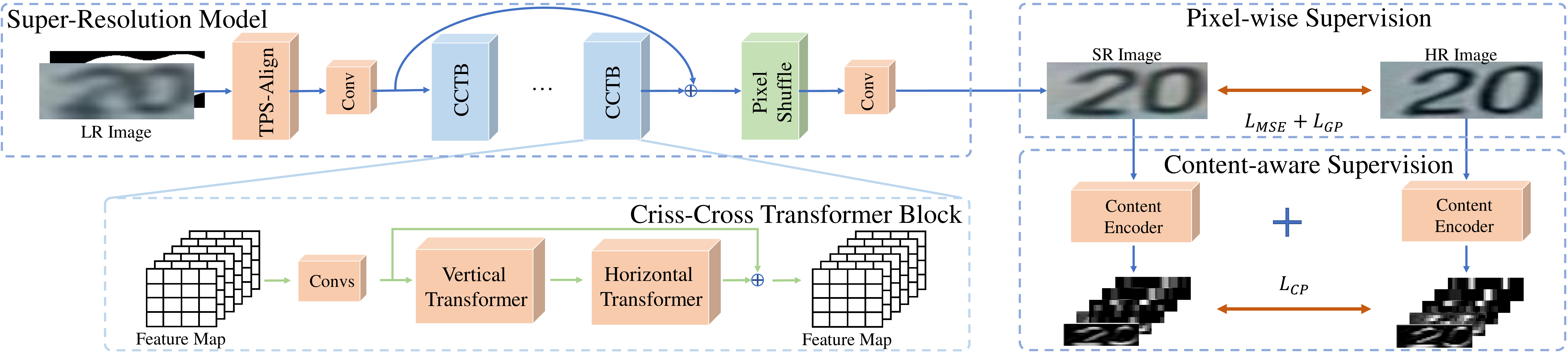}
	\caption{Overview of TATSR framework. Our method includes a super-resolution model, a pixel-wise supervision module, and a text-aware supervision module. The super-resolution model contains several consecutive CCTB blocks. The Content Encoder in Text-aware supervision is fixed during loss calculation.}
	\label{fig:pipeline}
\end{figure*}

\subsection{Overview\label{sec:pipeline}}
Our proposed TATSR framework is shown in Fig.~\ref{fig:pipeline}, which consists of three components: a super-resolution network, a pixel-level supervision module, and a text-aware supervision module. In our framework, the low-resolution~(LR) image and its binary mask are first aligned by the adaptive Thin-Plate-Spline (TPS,~\cite{stn}) module. Then, CNN features are extracted by a single convolution layer and go through a series of repeated Criss-Cross Transformer Blocks. Finally, we upsample the processed feature maps by the Pixel Shuffle~\cite{pixelshuffle} module to generate super-resolution results. The training of our network is supervised by the pixel-level supervision module and the text-aware supervision module. 
The pixel-wise supervision module works in RGB color space which directly calculates the L2 loss and Gradient Prior Loss~\cite{wang2020scene} with the high-resolution image. The text-aware supervision module works in text-oriented feature spaces to calculate the high-level similarity measured by our Content Perceptual Loss which integrates text information from multi-scale to overcome the shortcomings of the previous text-oriented losses.

\subsection{Content Perceptual Loss\label{sec:cploss}}
In this section, we introduce the Content Perceptual (CP) Loss in detail. Specifically, we extract the CNN part of the CRNN~\cite{crnn} model as the loss function network. The overall CRNN model has been pre-trained for scene text recognition, and the parameters are fixed during the loss calculation. The CNN part of the CRNN model has five downsampling operations through max-pooling and convolution. 
We extract the features after these five downsampling layers for loss calculation. Let $\phi_j(x),(j=1,2,3,4,5)$ be the activations after the $j$th downsampling layer of the loss function network $\phi$ when processing the input image $x \in R^{C_0 \times H_0 \times W_0}$. Assuming that the shape of $\phi_j(x)$ is $C_j \times H_j \times W_j$, the single-scale Content Perceptual Loss $\mathcal{L}_{\mathrm{fea}}^{j}$ after the $j$th downsampling layer between super-resolution (SR) image $I_S$ and high-resolution (HR) image $I_H$ can be calculated as follows:
\begin{equation}
	\mathcal{L}_{\mathrm{fea}}^{j}(\phi,I_S,I_H) = \frac{1}{C_jH_jW_j}||\phi_j(I_S)-\phi_j(I_H)  ||_2^2.
\end{equation}
The overall CP Loss consists of the weighted sum of each single-scale losses:
\begin{equation}
	\mathcal{L}_{\mathrm{CP}} (\phi,I_S,I_H)= \sum_{j=1}^{5} \lambda_j \cdot \mathcal{L}_{\mathrm{fea}}^{j}(\phi,I_S,I_H).
\end{equation}

As demonstrated in~\cite{johnson2016perceptual}, the shallow and deep features from the pre-trained CNN focus on the local structure and global semantics, respectively. Therefore, by simultaneously calculating the similarity between HR images and SR images on multi-scale Scene-Text-Recognition (STR)-trained features, CP Loss can ensure the consistency of high-level text contexts and low-level stroke structures at the same time. Since we use the distance between STR-based features as the similarity measurement instead of the final prediction results, we can avoid the convergence point conflict with pixel-wise losses caused by the direct loss calculation with text labels. Furthermore, unlike the recognition-result-based loss that strongly depends on the text strings, CP loss can transfer the text priors contained in the features to the non-training languages, which means that even if the recognition model is not trained on the specific language, it can also be used for CP loss calculation.

\begin{figure*}[htbp]
	\centering
	\includegraphics[width=\textwidth]{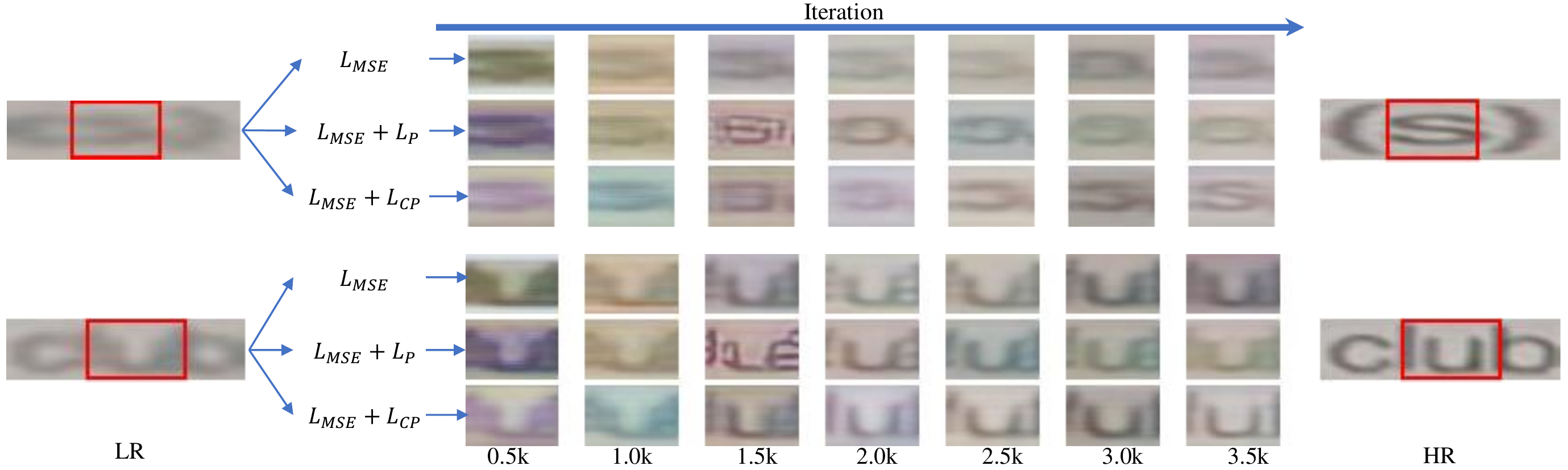}

	\caption{Comparison of the results of the same input image at different iterations. $L_\mathrm{MSE}$, $L_\mathrm{P}$, and $L_\mathrm{CP}$ denote the supervision of Mean Absolute Error Loss, Perceptual Loss, and Content Perceptual Loss, respectively. Note that the MSE loss alone produces the most blurry results. The perceptual loss can produce better results than MSE loss but not as good as the CP loss deal to the unique charactistics of text images.}
	\label{fig:iteration}

\end{figure*}

Similarly, the Perceptual Loss~\cite{johnson2016perceptual} in the Single Image Super-Resolution (SISR) task uses the measurement of image-classification-based features, achieving great performance in SISR. However, it should note that we have a clear difference in the scope of application. Because of the huge domain gap between natural and text images, the image-classification-trained features focus on the local details of general images, having a poor perception of the text semantics and character shapes. In contrast, the features obtained from pre-trained text recognition models contain more text-oriented information. They can better measure the similarity between the foreground characters in the SR and HR images. For a more intuitive explanation, we compare the iteration process of Mean Absolute Error (MSE) Loss, Perceptual Loss, and CP Loss in Fig.~\ref{fig:iteration}. Compared with MSE Loss's blurry results, Perceptual loss restores some unrecognizable local details due to its text-unsuitable general priors (\textit{e.g.,} 1.5k iters in Fig.~\ref{fig:iteration}) while CP Loss speeds up the optimization of text-oriented information, resulting in the quickest recovery of the important text foreground.

\noindent\textbf{Overall Loss Function} The overall loss function is composed of the pixel-wise part $\mathcal{L}_{\mathrm{PE}}$ and the content-aware part $\mathcal{L}_{\mathrm{CA}}$. The pixel-wise part is the weighted sum of L2 Loss $\mathcal{L}_{\mathrm{2}}$ and the  Gradient Prior Loss~\cite{wang2020scene} $\mathcal{L}_{\mathrm{GP}}$. The Content-aware part is our Content Perceptual (CP) Loss $\mathcal{L}_{\mathrm{CP}}$. The calculation of the overall loss can be represented as:
\begin{align}	\mathcal{L}_{\mathrm{PE}}&=\lambda_{\mathrm{2}}\mathcal{L}_{\mathrm{2}}+\lambda_{\mathrm{GP}}\mathcal{L}_{\mathrm{GP}},\\
\mathcal{L}_{\mathrm{CA}}& = \lambda_{\mathrm{CP}}\mathcal{L}_{\mathrm{CP}}, \\
\mathcal{L}&= \mathcal{L}_{\mathrm{PE}}+ \mathcal{L}_{\mathrm{CA}},  
\end{align}
where $\lambda_{\mathrm{2}}$, $\lambda_{\mathrm{GP}}$, and $\lambda_{\mathrm{CP}}$ are hyperparameters to trade off the proportions of three kinds of losses. In experiments, we fix $\lambda_{\mathrm{GP}}$ as $10^{-4}$, and further explore the balance of L2 Loss and CP Loss by changing the values of $\lambda_{\mathrm{2}}$ and $\lambda_{\mathrm{CP}}$. The experimental results can be found in supplementary, and we choose the best ($\lambda_{\mathrm{2}}=0.1, \lambda_{\mathrm{CP}}=5\times 10^{-4}$) as the final settings.

\subsection{Criss-Cross Transformer Block\label{sec:cctb}}
\begin{figure}[t]
	\centering
	\includegraphics[width=0.45\textwidth]{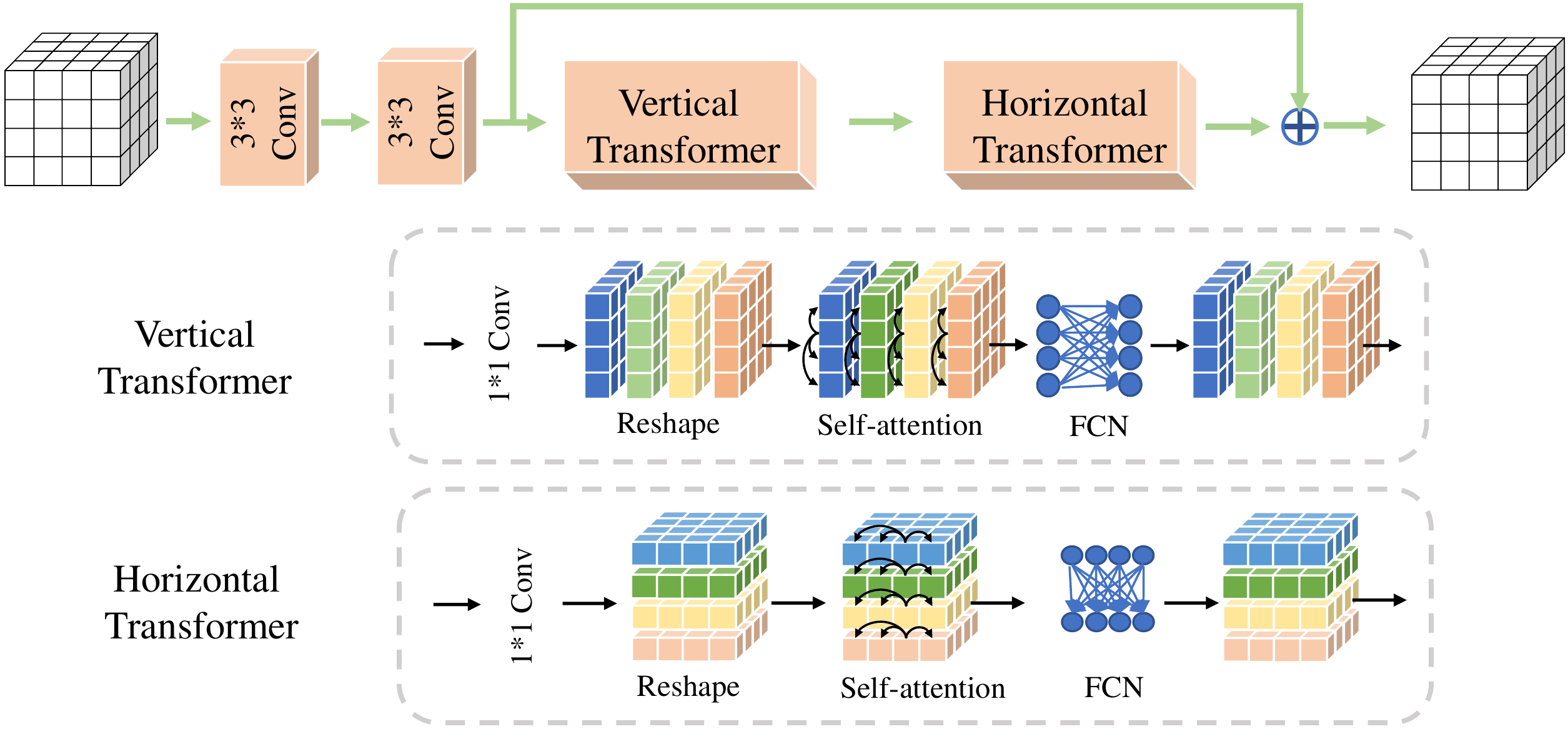}

	\caption{The structure of Criss-Cross Transformer Block. It consists of a vertical and a horizontal transfomer for long range information propagation. }
	\label{fig:CCTB}
\end{figure}

We detailed our Criss-Cross Transformer Block (CCTB) in this section. The specific structure of CCTB is shown in Fig.~\ref{fig:CCTB}.
CCTB contains two consequent transformer encoders, which we call vertical transformer $\phi_v$ and horizontal transformer $\phi_h$, responsible for co-column and co-row learning, respectively. Specifically, we denote $col^j(X)\in R^{C\times H \times 1}, row^i(X)\in R^{C\times 1 \times W}$ as the $j$th column and the $i$th row of the feature map $X\in R^{C\times H \times W}$. Assuming that the input feature map is $I$, the $i$th column of vertical transformer's output $O_v$ and the $j$th row of horizontal transformer's output $O_h$ can be calculated as:
\begin{align}
    &col^i(O_v) = \phi_{col}(col^i(I)),\\
    &row^j(O_h) = \phi_{row}(row^j(O_v)),    
\end{align}
The entire outputs of the vertical (horizontal) transformers are the concatenation of every single column (row) output,
\begin{align}
    &O_v = concat(col^1(O_v),col^2(O_v),col^3(O_v),...,col^W(O_v)),\\
    &O_h = concat(row^1(O_v),row^2(O_v),row^3(O_v),...,row^HO_v)).
\end{align}
This design splits the learning of different contexts into two orthogonal directions separately, which we call criss-cross learning. We choose the criss-cross strategy because most of the characters in natural scenes are arranged horizontally. Therefore, such a simple strategy can maximize the separation of text sequence and character shape information into the horizontal direction and vertical direction respectively. 
Intuitively, the learning of character shapes and text sequence information are mutually influenced. Better single-character information facilitates the learning of text sequences and vice versa. Thus, We choose the alternating layout strategy for the two kinds of transformers rather than the parallel one for a more frequent alternate optimization.
Compared to the previous ViT~\cite{vit}-like all-flattened image-to-sequence strategy in STT~\cite{scenetexttelescope}, this explicit text context separation strategy reduces the learning difficulty of each text context, effectively boosting the transformer's performance in fitting different text context. 

The global visibility of multi-head attention makes CCTB can obtain the complete sequence in the horizontal direction. It is able to memorize whole sequence information equally without forgetting any middle units. Therefore, CCTB can accurately model the sequence contexts of samples with arbitrary length texts  without the problem of RNN's long-time memory failure, leading to better-predicting character types and fewer restored characters with type error. This is especially beneficial for long text sample processing which is more dependent on sequence information.

\section{Experiment\label{sec:Experiment}}
\subsection{Datasets and Metrics}

\subsubsection{Datasets}
 \noindent \textbf{TextZoom}~\cite{wang2020scene} dataset is a real-scene text image dataset cropped from RealSR~\cite{cai2019toward} and SRRAW~\cite{zhang2019zoom}. 
 TextZoom contains $17,367$ LR-HR image data pairs for training and three test subsets divided according to the focal lengths when shot, namely easy ($1,619$ samples), medium ($1,411$ samples), and hard ($1,343$ samples).

\noindent \textbf{ChineseSTR} is a scene text image dataset with Chinese texts as the main content generated by us, consisting of $161,219$ train samples and $8,508$ test samples. We sample high-resolution images from PaddleOCR Datasets and synthesize the low-resolution version by the compound degradations in BSRGAN~\cite{bsrgan}.

\begin{table*}[htbp]
	\begin{center}
		\caption{Performance of the mainstream scene text image super-resolution methods on the three validation sets of TextZoom~\cite{wang2020scene}. $L_{\mathrm{TV}}$ denotes Total Variation Loss. $L_{\mathrm{P}}$ denotes Perceptual Loss~\cite{johnson2016perceptual}. $L_\mathrm{GP}$ denotes Gradient Prior Loss~\cite{wang2020scene}. $L_{\mathrm{C/PA}}$ denotes the Content-aware/Position-aware Loss~\cite{scenetexttelescope}. $L_{\mathrm{CP}}$ denotes our Content Perceptual Loss.}\vspace{-0.2in}
		\label{table:SOTA_Cmp}
		\resizebox{\textwidth}{28mm}{
			\begin{tabular}{cc|cccc|cccc|cccc}
				\toprule[1.0pt]
				\multirow{2}*{Method} & \multirow{2}*{Loss} & \multicolumn{4}{|c|}{Aster~\cite{shi2018aster}} & \multicolumn{4}{|c|}{Moran~\cite{luo2019moran}} & \multicolumn{4}{|c}{CRNN~\cite{crnn}}\\
				
				~ & ~ & easy & medium & hard & all & easy & medium & hard & all & easy & medium & hard & all \\
				\midrule[0.5pt]
				BICUBIC & - & 64.7\% & 42.4\% & 31.2\% &47.2\% & 60.6\%& 37.9\% & 30.8\% &44.1\% & 36.4\% & 21.1\% & 21.1\% & 26.8\% \\ 
				
				\midrule[0.5pt]
				SRCNN~\cite{srcnn} & $L_{2}$ & 69.4\% &43.4\% &32.2\% &49.5\% & 63.2\% & 39.0\% & 30.2\% & 45.3\% & 38.7\% & 21.6\% & 20.9\% & 27.7\% \\
				
				
				SRResNet~\cite{srres} &$L_{2}+L_{\mathrm{TV}} +L_{\mathrm{P}}$ &69.4\%& 47.3\% & 34.3\% & 51.3\% & 60.7\% &42.9\% &32.6\% &46.3\% &39.7\% & 27.6\% & 22.7\% & 30.6\%\\
				
				RRDB~\cite{RRDB} & $L_{1}$ & 70.9\% & 44.4\% &32.5\% &50.6\% &63.9\% & 41.0\% &30.8\% &46.3\% &40.6\% &22.1\% &21.9\% & 28.9\% \\
				
				EDSR~\cite{EDSR} & $L_{1}$ & 72.3\% & 48.6\% &34.3\% &53.0\% &63.6\% & 45.4\% &32.2\% &48.1\% &42.7\% &29.3\% &24.1\% & 32.7\% \\
				
				LapSRN~\cite{LapSRN} & $L_{\mathrm{charbonnier}}$ & 71.5\% & 48.6\% &35.2\% &53.0\% &64.6\% & 44.9\% &32.2\% &48.3\% &46.1\% &27.9\% &23.6\% & 33.3\% \\
				
				\midrule[0.5pt]
				TSRN~\cite{wang2020scene} & $L_{2}+L_{\mathrm{GP}}$ & 75.1\% & 56.3\% & 40.1\% & 58.3\% & 70.1\% & 53.3\% & 37.9\% & 54.8\% & 52.5\% & 38.2\% & 31.4\% & 41.4\% \\
				
				STT~\cite{scenetexttelescope} & $L_{2}+L_{\mathrm{C/PA}}$ & 75.7\% & 59.9\% & 41.6\% & 60.1\% & 74.1\% & 57.0\% & 40.8\% & 58.4\% & 59.6\% & 47.1\% & 35.3\% & 48.1\%\\
				
				PCAN~\cite{2021Parallelly} & $L_{2}+L_{\mathrm{EG}}$ & 77.5\% & 60.7\% & 43.1\% & 61.5\% & 73.7\% & 57.6\% & 41.0\% & 58.5\% & 59.6\% & 45.4\% & 34.8\% & 47.4\% \\
				TG~\cite{TG} & $L_{2}+L_{\mathrm{SFM}}$ & 77.9\%	& 60.2\%	& 42.4\%	& 61.3\%	& 75.8\%	& 57.8\%	& 41.4\%	& 59.4\%	& 61.2\%	& 47.6\%	& 35.5\%	& 48.9\% \\
				TATT~\cite{TATT}& $L_{2}+L_{\mathrm{TP}}+L_{\mathrm{TSC}}$  & 78.9\%	& 63.4\%	& 45.4\%	& 63.6\%	& 72.5\%	& 60.2\%	& 43.1\%	& 59.5\%	& 62.6\%	& 53.4\%	& 39.8\%	& 52.6\% \\
				C3-STISR~\cite{C3-STISR} & $L_{2}+L_{\mathrm{C/PA}}+L_{\mathrm{rec}}+L_{\mathrm{ling}}$ & 79.1\%	& 63.3\%	& \textbf{46.8}\%	& 64.1\%	& 74.2\%	& \textbf{61.0}\%	& 43.2\%	& 60.5\%	& 65.2\%	& 53.6\%	& 39.8\%	& 53.7\% \\

				\midrule[0.5pt]
				TATSR(ours) & $L_{2}+L_{\mathrm{GP}}+L_{\mathrm{CP}}$ &\textbf{ 80.4}\% & \textbf{64.6}\% & 46.0\% & \textbf{64.7}\% & \textbf{75.4}\% & 60.7\% & \textbf{44.1}\%	& \textbf{61.0}\% & \textbf{66.8}\% & \textbf{54.2}\% & \textbf{40.1}\% & \textbf{54.2}\% \\
				\bottomrule[1.0pt]
			\end{tabular}
		}
	\end{center}
\end{table*}

\subsubsection{Evaluation Metrics}
Since the primary purpose of this task is to improve the text recognition accuracy and human eye perception of text images, we choose the recognition accuracy of the open-source text recognition models and visual perception to measure the model's performance. Specifically, we use the text recognition results of the pre-trained Aster~\cite{shi2018aster}, CRNN~\cite{crnn}, and Moran~\cite{luo2019moran} models for evaluation. To be consistent with the previous work, we exclude the influence of punctuation and capitalization when calculating the accuracy. 
\begin{figure}[h]
	\includegraphics[width=0.475\textwidth]{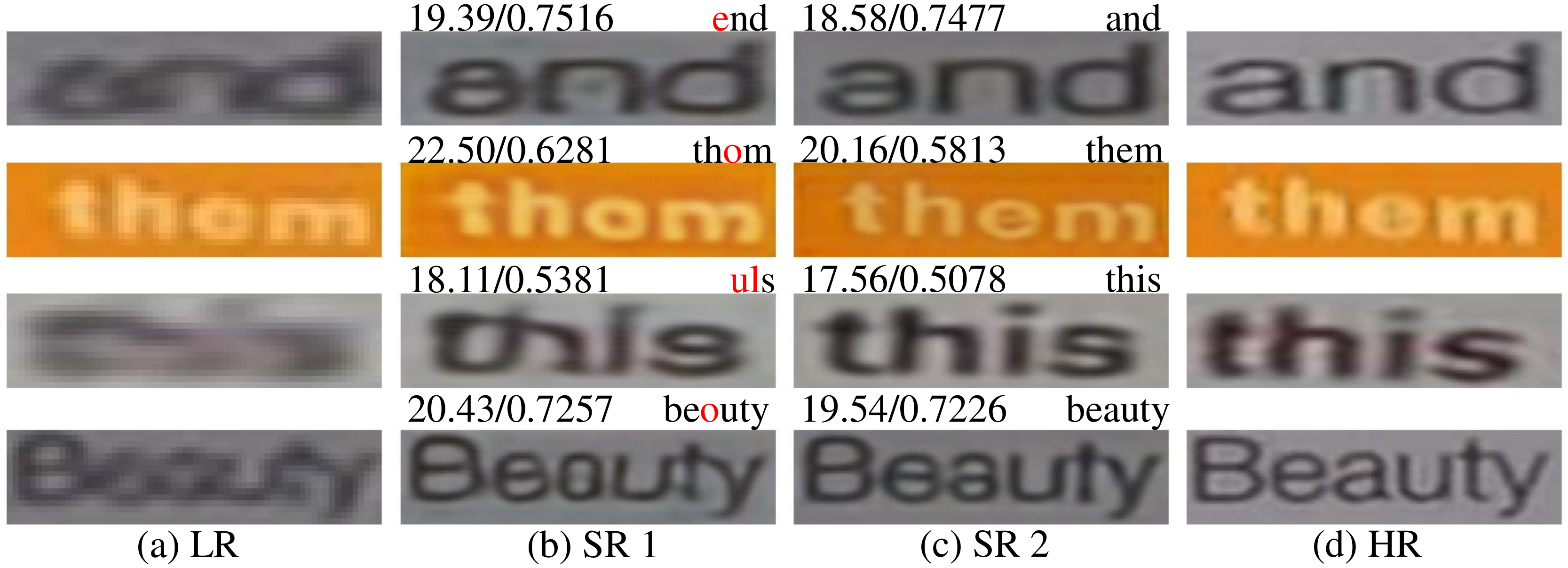}
	\caption{Examples of conflicts between PSNR/SSIM and text recognition accuracy/visual perception. The labels above represent PSNR, SSIM, and CRNN~\cite{crnn}'s results. Note the results with higher PSNR/SSIM does not correspond to the results with higher recognition accuracy.} 
	\label{fig:psnr_ssim_err}

\end{figure}

Similar to previous work~\cite{scenetexttelescope,TG,2021Parallelly,TATT}, we do not use Peak Signal-to-Noise Ratio (PSNR) and Structure Similarity Index Measure (SSIM) as the main evaluation metrics because they are not suitable for text image evaluation. In text images, the character regions with a small proportion of pixels have a far more significant impact on perception than the large background. However, as mentioned in previous work~\cite{scenetexttelescope}, these metrics are averaged based on all image pixels, causing a much larger divergence than in the general images between the metrics and human perception. More intuitively, a restored character with sharp but misaligned boundaries (\textit{e.g.,} Fig.~\ref{fig:psnr_ssim_err} (c)) tends to have lower PSNR/SSIM than a restored character with a blurry appearance (\textit{e.g.,} Fig.~\ref{fig:psnr_ssim_err} (b)). This problem is more obvious on artificially cropped real scene datasets (\textit{e.g.,} TextZoom). Therefore, the visual perception and text recognition accuracy are more accurate metrics for judging the performance of scene text image super-resolution.

\subsubsection{Implementation Details}
Our method is implemented using the PyTorch framework. For data preprocessing, following the previous works, we resize the low-resolution images and high-resolution images to $16\times 64$ and $32\times 128$, respectively. All experimental conditions are trained on $5$ NVIDIA RTX $2080$ Ti GPUs. Unless otherwise stated, we set the number of CCTBs and feature channels to 4 and 128, respectively. The network is optimized using the Adam optimizer, and the initial learning rate is set to $5 \times 10^{-4}$. Training takes $500$ epochs, and the batch size is set to $128$. The models used for text recognition are based on the officially released codes and the pre-trained parameters.

\subsection{Comparison with State-of-the-Art}
\begin{figure*}[h]
	\centering
	\includegraphics[width=\textwidth]{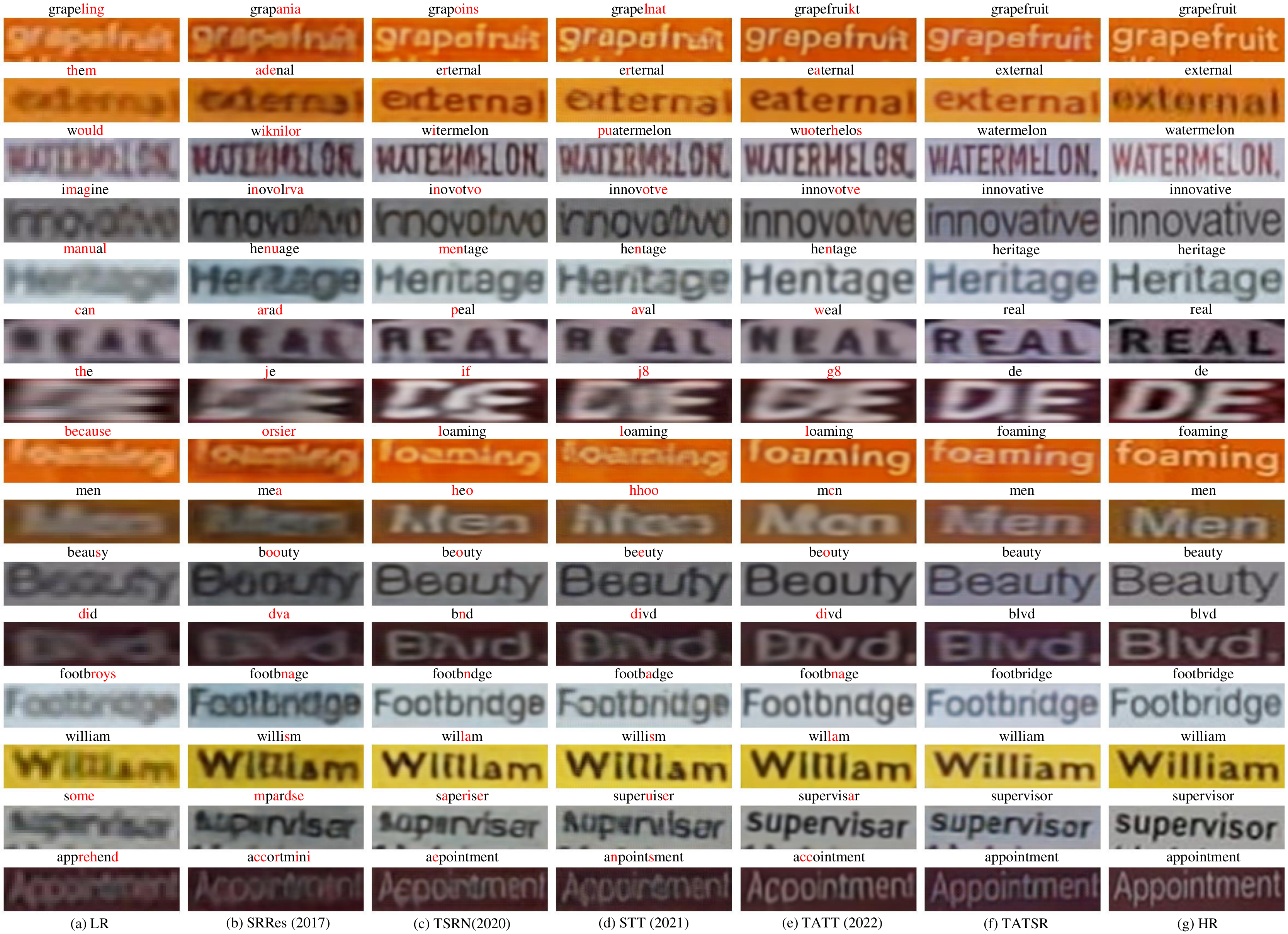}
	\caption{Examples of the restored images in TextZoom and their recognition results of CRNN~\cite{crnn}. Our method not only produces the clearest results but also achieve the best performance in text recognition after restoration.}
	\label{fig:sota_cmp}
\end{figure*}

In Tab.~\ref{table:SOTA_Cmp} and Fig.~\ref{fig:sota_cmp}, we list the quantitative and qualitative results of twelve different competitive methods on the TextZoom~\cite{wang2020scene}, including SRCNN~\cite{srcnn}, SRResNet~\cite{srres}, RRDB~\cite{RRDB}, EDSR~\cite{EDSR}, LapSRN~\cite{LapSRN} belonging to the Single Image Super-Resolution methods and TSRN~\cite{wang2020scene}, STT~\cite{scenetexttelescope}, PCAN~\cite{2021Parallelly}, TG~\cite{TG}, TATT~\cite{TATT}, C3-STISR~\cite{C3-STISR} and TATSR belonging to Scene Text Image Super-Resolution methods. 

First, the quantitative results in Tab.~\ref{table:SOTA_Cmp} show that TATSR significantly improves the text recognition accuracy under the conditions of every text recognition model and test subset. Compared with BICUBIC, TATSR achieves $17.5\%$, $16.9\%$, and $27.4\%$ improvement under the three text recognition models, respectively. Furthermore, it also reaches the best recognition accuracy under all three different recognition models, proving that TATSR effectively enhances the text recognition quality of low-resolution images. 

Second, compared with visualization of other methods in Fig.~\ref{fig:sota_cmp}, TATSR can generate more regular and precise strokes and more readable characters by using our Content Perceptual (CP) Loss. Besides, the Criss-Cross Transformer Blocks (CCTBs) can better handle the text characteristics, especially when dealing with images containing long character sequences (\textit{e.g.,} row 3,4,12 and 15 in Fig.~\ref{fig:sota_cmp}). 
In general, TATSR has achieved consistent improvements in both human eye perception and text recognition compared to the previous methods.

\begin{table}[htbp]
		\centering
		\caption{Aster~\cite{shi2018aster} recognition accuracy of different models trained with and without Content Perceptual Loss.}
		\label{table:CPEffect_ablation}
			\begin{tabular}{c|c|cccc}
				\toprule[1.0pt]
				Method & $L_{CP}$  &  easy & medium & hard & all \\
				\midrule[0.5pt]
				\multirow{2}*{SRCNN~\cite{srcnn}} & $ \times$ &  69.4\% &43.4\% &32.2\% &49.5\% \\
				~ & $ \checkmark$ & \textbf{70.5}\% & \textbf{45.7}\% & \textbf{33.6}\% & \textbf{51.2}\% \\
				\midrule[0.5pt]
				\multirow{2}*{SRResNet~\cite{srres}}& $ \times$ & 69.4\%& 47.3\% & 34.3\% & 51.3\% \\
				~ & $ \checkmark$ & \textbf{73.0}\% & \textbf{56.4}\% & \textbf{39.7}\% & \textbf{57.4}\%  \\
				\midrule[0.5pt]
				\multirow{2}*{TSRN~\cite{wang2020scene}} & $ \times$ & \textbf{75.1}\% & 56.3\% & 40.1\% & 58.3\% \\
				~ & $ \checkmark$ & 74.9\%&	\textbf{60.6}\%	& \textbf{42.1}\%	&\textbf{60.2}\% \\
				\midrule[0.5pt]
				 \multirow{2}*{TBSRN~\cite{scenetexttelescope}} & $ \times$ & 75.2\% & 56.7\% & 40.2\% & 58.5\% \\
				 ~ & $ \checkmark$ & \textbf{77.0}\%&	\textbf{58.5}\%	& \textbf{42.7}\%	& \textbf{60.5}\% \\
				 \midrule[0.5pt]
				\multirow{2}*{TATSR}  & $ \times$  & 75.6\%	& 58.9\%	&42.7\%	& 60.1\%\\
				~  & $ \checkmark$  &\textbf{ 80.4}\% & \textbf{64.6}\% & \textbf{45.9}\% & \textbf{64.7}\%\\
				\bottomrule[1.0pt]
			\end{tabular}
	\end{table}
	\begin{table}[t]
		\centering
		\caption{Aster~\cite{shi2018aster} recognition accuracy of models trained with different text-oriented losses. '$\times$' denotes without the supervision of any of these losses.}
		\label{table:CPad_ablation}
			\begin{tabular}{c|c|cccc}
				\toprule[1.0pt]
				Method & Loss  & easy & medium & hard & all \\
				\midrule[0.5pt]
				\multirow{4}*{TSRN~\cite{wang2020scene}} & $ \times $&  75.1\% & 56.3\% & 40.1\% & 58.3\% \\
				~ & $ +L_{p} $&  \textbf{75.4\%} & 58.3\% & 41.7\% & 59.5\% \\
				~ & $ +L_{C/PA} $ & 74.3\% & 59.7\% & 39.6\% & 58.9\% \\
				~ & $+L_{CP}$& 74.9\%&	\textbf{60.6}\%	& \textbf{42.1}\% &	\textbf{60.2}\% \\
                \midrule[0.5pt]
				\multirow{4}*{TBSRN~\cite{scenetexttelescope}} & $ \times $ & 75.2\% & 56.7\% & 40.2\% & 58.5\% \\
				~ & $ +L_{p} $&  73.4\% & 56.6\% & 41.6\% & 58.2\% \\
				~ & $ +L_{C/PA} $ & 75.7\% & \textbf{59.9}\% & 41.6\% & 60.1\% \\
				~ & $+L_{CP}$& \textbf{77.0}\% &	58.5\%	& \textbf{42.7}\%& \textbf{60.5}\% \\
				\midrule[0.5pt]
				\multirow{4}*{TATSR} & $ \times $&  75.6\%	& 58.9\%	&42.7\%	& 60.1\%\\
				~ & $ +L_{p} $&  76.0\% & 62.6\% & 45.0\% & 62.1\% \\
				~ & $ +L_{C/PA} $ & 76.9\% & 63.0\% & 44.0\% & 62.6\% \\
				~ & $+L_{CP}$&\textbf{ 80.4}\% & \textbf{64.6}\% & \textbf{45.9}\% & \textbf{64.7}\%\\

				\bottomrule[1.0pt]
			\end{tabular}
	\end{table}

\begin{figure*}[h]
	\includegraphics[width=\textwidth]{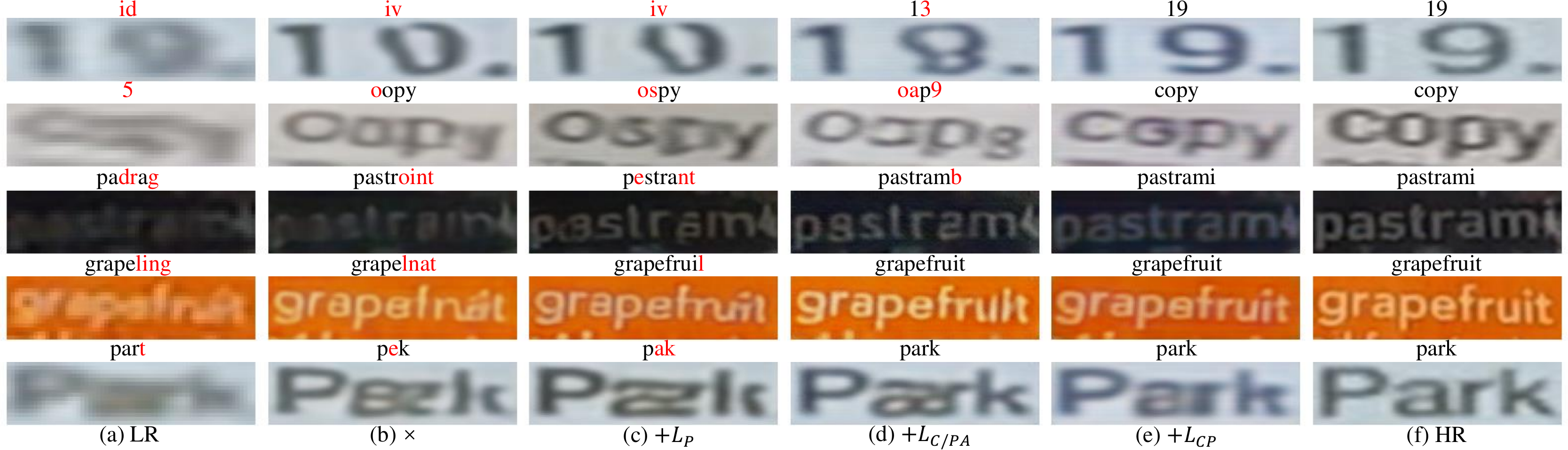}
	\caption{The super-resolution results of different text-oriented losses and the recognition results of CRNN~\cite{crnn}.} 
	\label{fig:loss_cmp}
\end{figure*}

\subsection{Ablation Study\label{sec:ablation}}

\subsubsection{Effectiveness of Content Perceptual (CP) Loss}
To verify the effectiveness of CP Loss, we used it to train five different models, and Tab.~\ref{table:CPEffect_ablation} shows the results of each model trained with and without CP Loss. Since our CP Loss uses the pre-trained CRNN~\cite{crnn} features, for fairness, we don't use the recognition accuracy of CRNN~\cite{crnn} as the evaluation methods for the comparison of CP Loss and other text-oriented losses. Instead, we choose the Aster~\cite{shi2018aster} recognition accuracy as the evaluation metrics in the ablation study of CP Loss. The factor of text-oriented loss will also be blocked in the comparison of CCTB with other blocks (Tab.~\ref{table:CCTB_ablation} and Fig.~\ref{fig:block_cmp}). CP Loss brings obvious improvement to each model, and it is worth noting that with CP Loss, SRResNet~\cite{srres} designed for the single image super-resolution can achieve competitive precision to TSRN~\cite{wang2020scene} specially designed for scene text image super-resolution, showing CP Loss's great boost.

Moreover, we compare the performance of Perceptual (P) Loss~\cite{johnson2016perceptual}, Content/Position Aware(C/PA) Loss~\cite{scenetexttelescope}, and our CP Loss by training the same models. The quantitative and qualitative results are shown in Tab.~\ref{table:CPad_ablation} and Fig.~\ref{fig:loss_cmp}, which indicate the advantage of our CP loss over the other text-oriented losses. 
Compared to text-imperceptible Perceptual Loss, text-oriented C/PA Loss and CP Loss can provide text-aware supervision, effectively reducing misjudgment of character types and getting higher recognition accuracy. Since it also supervises the low-level features while providing deep text-aware supervision, our CP Loss can recover finer and sharper strokes than C/PA Loss, further improving the visual perception and reducing the recognition difficulty.
In addition, the improvement brought by all the auxiliary losses, including CP Loss, are more obvious on TATSR than on TSRN~\cite{wang2020scene} and TBSRN~\cite{scenetexttelescope}, showing that while increasing the text relevance of supervision, the improvement of the model's corresponding learning ability is also necessary.

\begin{table}[tbp]
	\centering
	\caption{Comparison of the proportion of different single-layer losses $L_{fea}^j$ in $L_{CP}$.}
	\label{table:Hypara_ablation2}
		\begin{tabular}{c|cccc}
			\toprule[1.0pt]				
			$ \lambda_1: \lambda_2: \lambda_3: \lambda_4: \lambda_5$ &  easy & medium & hard & all \\
			\midrule[0.5pt]
			$ 1.6:1.6:1.6:0.0:0.0$ & 78.0\%	&63.9\%	&45.6\%	&63.5\% \\
			$1.4:1.4: 1.4:0.4:0.4$ &\textbf{ 80.4\%} & \textbf{64.6\%} & \textbf{45.9\%} & \textbf{64.7\%}\\
			$1.0:1.0: 1.0:1.0:1.0$ &79.1\%	&63.4\%	&47.1\%	&64.2\%\\
			$0.5:0.9: 1.2:1.2:1.2$ &76.2\%	&61.9\%	&45.7\%	&62.2\% \\
			
			\bottomrule[1.0pt]
		\end{tabular}
\end{table}	
\begin{table}[tbp]
	\centering
	\makeatletter\def\@captype{table}\makeatother\caption{Comparison of the performance of TATSR with different assignments to $L_{MSE}$, $L_{GP}$, and $L_{CP}$.}
	\label{table:Hypara_ablation}
		\begin{tabular}{c|cccc}
			\toprule[1.0pt]				
			$ \lambda_{\mathrm{MSE}}: \lambda_{\mathrm{GP}}: \lambda_{\mathrm{CP}}$   & easy & medium & hard & all \\
			\midrule[0.5pt]
			$ 1.0:0.0001:0.0000$ & 75.6\% & 58.9\% & 42.7\% & 60.1\% \\
			$1.0:0.0001: 0.0001$ & 76.5\%	&60.5\%	&44.8\%	&61.6\% \\
			$1.0:0.0001: 0.0005$ & 76.6\%	&61.6\%	&42.4\%	&61.3\% \\
			$0.1:0.0001: 0.0005$ &\textbf{ 80.4\%} & \textbf{64.6\%} & \textbf{45.9\%} & \textbf{64.7\%}\\
			
			\bottomrule[1.0pt]
		\end{tabular}
	
\end{table}

\subsubsection{Interpretability Analysis of Content Perceptual (CP) Loss}

\begin{figure*}[t]  
	\includegraphics[width=1\textwidth]{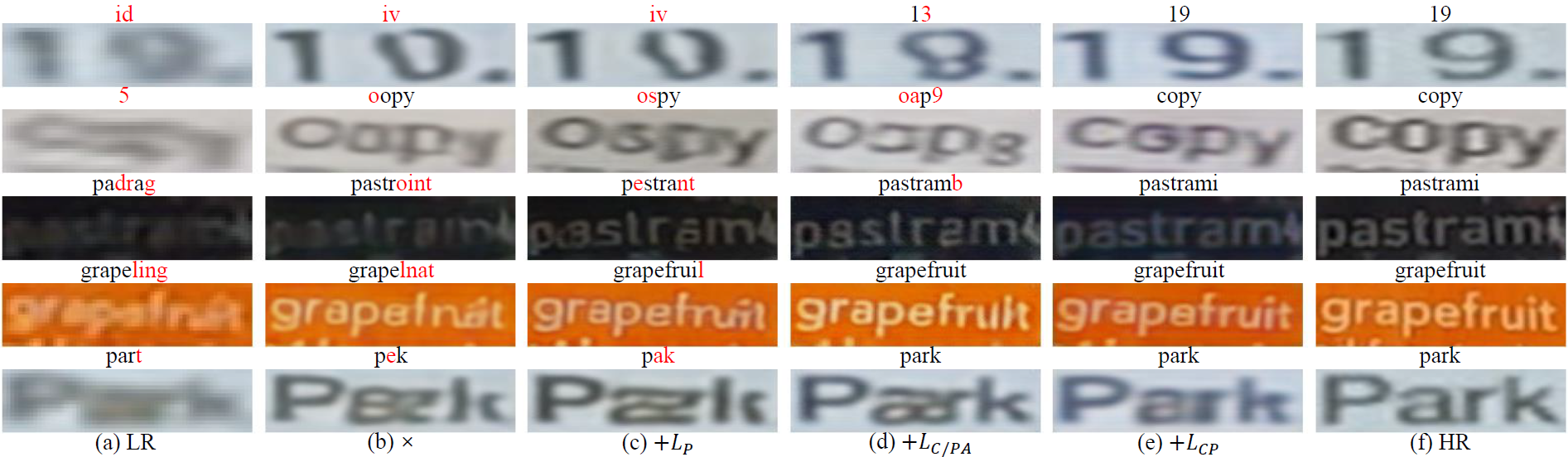}
	\caption{Examples of TATSR's results supervised by one single-layer loss $L_{fea}^{i}$ from the $j$th activations $\phi_i$. } \vspace{-0.2in}
	\label{fig:single layers}
\end{figure*}

For further analysis of the CP Loss, we conduct ablative experiments on feature scales and weight distribution. The results can be found in Tabs.~\ref{table:Hypara_ablation2},~\ref{table:Hypara_ablation} and Fig.~\ref{fig:single layers}. As shown in Fig.~\ref{fig:single layers}, the features extracted from deeper layers tend to contain more global semantics such as character contours and stroke layouts but less spatial details such as stroke shapes and edges, leading to contour-recognizable characters but blurry strokes. 
In contrast, the lower layers' features tend to generate sharper strokes but low recognizability of overall characters, caused by their sensitivity to low-level stroke shape and edge information but the limited perception of high-level semantics.
Tab.~\ref{table:Hypara_ablation2} shows that combining low-level structure and global semantics can bring the best performance, and the excessive focus on local structure or global semantics can lead to the loss of the other kind of information. 
We choose the weight proportion with the best result as the final setting of CP Loss.

In addition, we explore different proportions between pixel-wise losses and CP Loss. The results in Tab.~\ref{table:Hypara_ablation} show that by increasing the weight of CP Loss, the network focuses more on minimizing the distance between HR and SR images in text-aware feature spaces constructed by the multi-scale recognition-based features and pays less attention to the color space, leading to better-restored text foreground and a higher text recognition accuracy. It is necessary to explain that since the input of CRNN is a grayscale image, just using CP Loss and discarding pixel level loss will lead to instability in the learning process, leading to training failure. In general, we choose the settings of the best performance as the final loss settings.

\subsubsection{Cross-language Generalization of Content Perceptual (CP) Loss}
\begin{table}[t]
	\centering
	\caption{Multi-language Experiment.}
	\label{table:Multi_language}
		\begin{tabular}{c|cc}
			\toprule[1.0pt]
			Loss  & TextZoom(English) & ChineseSTR(Chinese)\\
			\midrule[0.5pt]
			$\times $ &  60.1\% & 40.8\% \\
			$+L_{P}$ &  62.1\% & 41.1\% \\
			$+L_{CP_{Chn}}$ &  62.5\% & \textbf{42.3}\%\\
			$+L_{CP_{Eng}}$ &  \textbf{64.7\%} &  41.9\% \\
			\bottomrule[1.0pt]
		\end{tabular}

\end{table}

To verify the cross-language generalization performance of CP Loss, we conducted cross-language comparative experiments on the Chinese-main dataset ChineseSTR and the English dataset TextZoom. Since CP Loss depends on the pre-trained text recognition model on the corresponding language, we also train a CRNN~\cite{crnn} model on ChineseSTR for Chinese-based CP Loss calculation.

\begin{figure}[t]

	\includegraphics[width=0.49\textwidth]{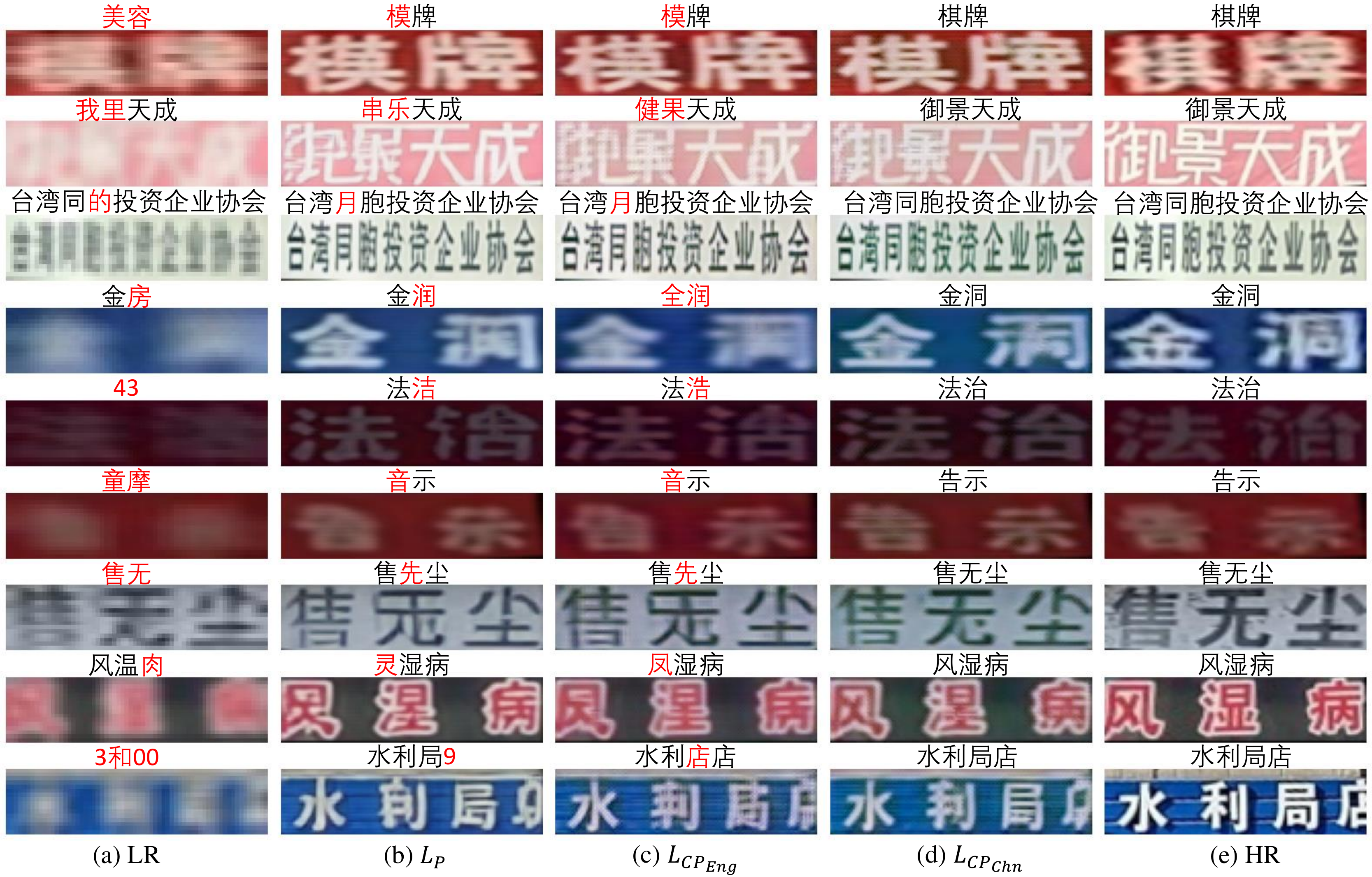}
	\caption{The examples of super-resolution results on ChineseSTR and their recognition results of Chinese-trained CRNN~\cite{crnn}.} 
	\label{fig:loss_cmp_ch}

\end{figure}

As shown in Tab.~\ref{table:Multi_language}, the losses for the corresponding language in both datasets achieve the best results, showing the excellent generalization performance of CP Loss in different languages. In addition, CP Loss also achieves the second best results on cross-lingual datasets, better than Perceptual Loss. 
This is consistent with the visualization in Fig.~\ref{fig:loss_cmp_ch}. Compared with Perceptual Loss, the Chinese stroke shapes recovered by English-based CP Loss are more recognizable. However, since English-based CP Loss does not contain Chinese character sequence information, it is more prone to misjudge Chinese character types than the Chinese-based CP Loss (\textit{e.g.,} row 3 in Fig.~\ref{fig:loss_cmp_ch}). Such results validate two points: 1)  CP Loss contains plenty of text priors including language-type specific and cross-language-available priors. CP loss of specific language gets the best performance on the exact language. 2) Texts across languages also have stronger domain priors (\textit{e.g.,} stroke shapes) than the generic image priors and such priors can work on cross-language cases. Our CP Loss can transfer these priors to text images in different languages, getting better results than the generic image prior loss.

\subsubsection{Effectiveness of CCTB}
\begin{figure*}[tbp]

	\includegraphics[width=\textwidth]{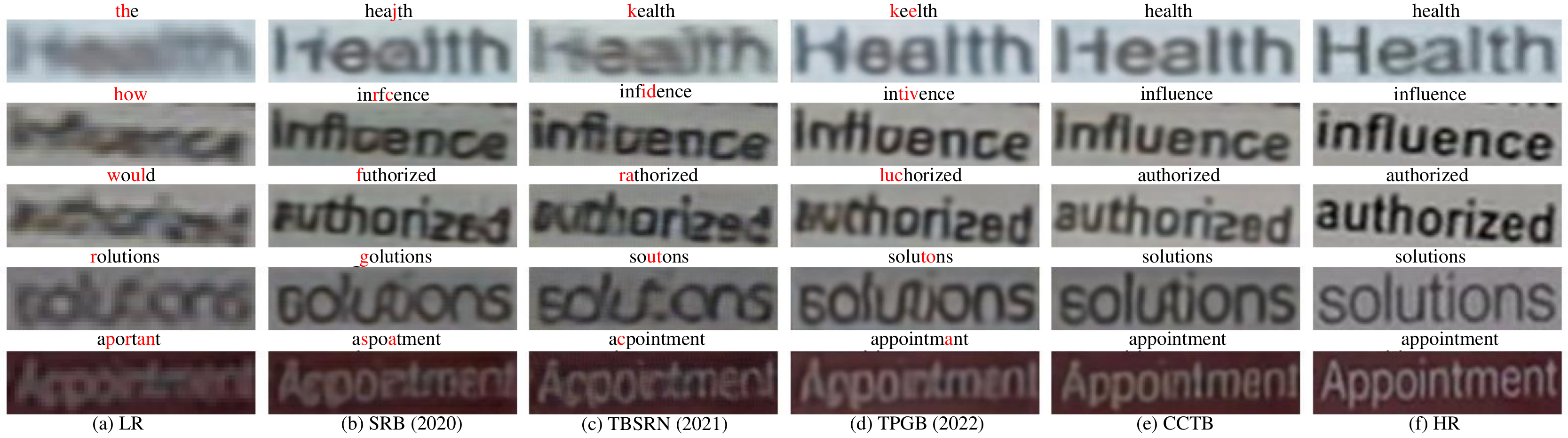}

	\caption{Comparison of super-resolution results of samples with long text sequences by different blocks.} 
    \vspace{-0.2in}
	\label{fig:block_cmp}

\end{figure*}
\begin{table}[tbp]
	\begin{center}
		\caption{Aster~\cite{shi2018aster} accuracy of super-resolution results generated by different sequence processing blocks.}

		\label{table:CCTB_ablation}
			\begin{tabular}{c|cccc}
				\toprule[1.0pt]
				Sequence Block & easy & medium & hard & all \\
				\midrule[0.5pt]
				SRB~\cite{wang2020scene} & 75.1\% & 56.3\% & 40.1\% & 58.3\% \\
				\midrule[0.5pt]
				TBSRN~\cite{scenetexttelescope} & 75.2\% & 56.7\% & 40.2\% & 58.5\%\\
				\midrule[0.5pt]
				TPGB~\cite{TATT} & 74.7\% & 57.8\% & \textbf{41.4}\% & 58.8\%\\
				\midrule[0.5pt]
				CCTB& \textbf{76.8}\%	& \textbf{61.0}\%	&41.3\%	& \textbf{60.8}\%\\
				
				\bottomrule[1.0pt]
			\end{tabular}
	\end{center}
\vspace{-0.2in}
\end{table}

To verify the effectiveness of our Criss-Cross Transformer Block (CCTB), 
 we choose Sequential Residual Block (SRB,~\cite{wang2020scene}), Transformer-Based Super-Resolution Network (TBSRN,~\cite{scenetexttelescope}), Text-Prior Guided Blocks (TPGB,~\cite{TATT}) and our CCTB for comparison. For fairness, we compare all sequence modules under the same frameworks and using the same supervision for training. Specifically, we use the vanilla TSRN~\cite{wang2020scene} framework supervised by L2 Loss and Gradient Prior Loss~\cite{wang2020scene} as the unified training framework and fix the number of blocks and feature channel to commonly used $5$ and $64$, respectively. The quantitative results in Tab.~\ref{table:CCTB_ablation} validate our CCTB's general superiority over the other blocks.

\begin{figure}[t]
        \centering
	\includegraphics[width=0.49\textwidth]{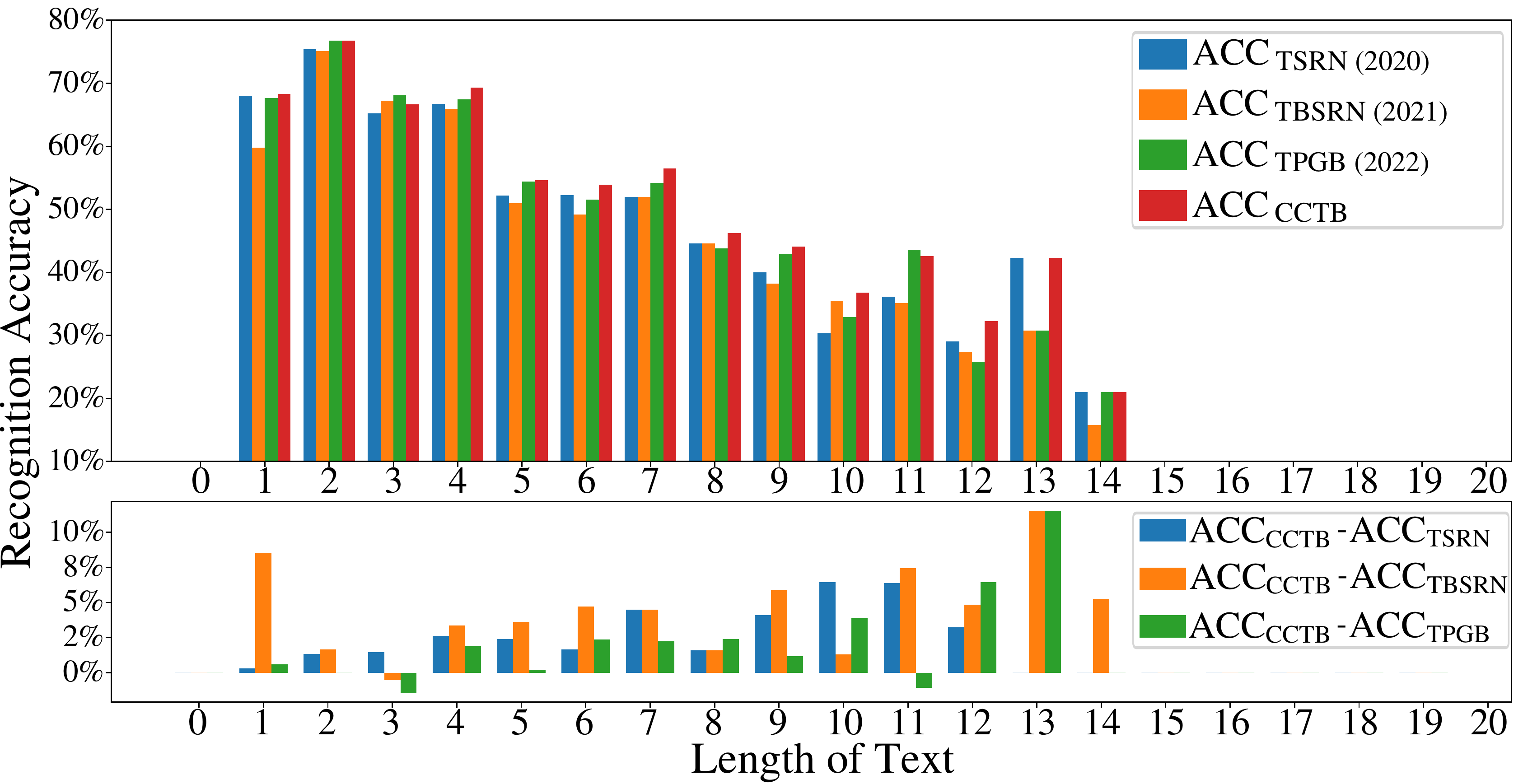}
	\caption{Top:  Different blocks' recognition accuracy on samples with different-length texts; Bottom: CCTB's recognition accuracy improvement over the other blocks on samples with different-length texts.} 
	\label{fig:block_acc_dis}

\end{figure}

For a detailed comparison,  we display the visualization of different blocks' results on long text samples in Fig.~\ref{fig:block_cmp} and CCTB's recognition accuracy improvement over the other blocks on samples with different-length texts in Fig.~\ref{fig:block_acc_dis}. When dealing with long text images (Fig.~\ref{fig:block_cmp}), RNN-based SRB and TPGB cannot accurately infer specific character types from the context information due to RNN's poor ability to model long-distance relationships. 
Though transformer-based TBSRN generates fewer recognition errors, its results' character shape is still very rough and hardly recognized by human perception. In contrast, our CCTB is able to adequately learn sequence information and character shape by two text-layout-oriented orthogonal transformers, respectively. 
Fig.~\ref{fig:block_acc_dis} shows the average recognition accuracy against text length, CCTB outperforms the other blocks on almost all samples with different text length.

\begin{table}[t]
	\begin{center}

		\caption{Aster~\cite{shi2018aster} accuracy of samples with different lengths of text. $\mathrm{Len}$ denotes the text length.}

		\label{table:CCTB_ablation_2}
			\begin{tabular}{cc|ccc}
				\toprule[1.0pt]
			    Hon Transformer & Ver Transformer & $\mathrm{Len}<9$ & $\mathrm{Len}>=9$ & all \\
				\midrule[0.5pt]
				$\times$ & $\times$   & 64.6\% & 34.6\% & 60.7\%\\
				\midrule[0.5pt]
				$\checkmark$ & $\times$  & 65.7\% & 37.6\% &  62.0\% \\
				\midrule[0.5pt]
				$\checkmark$ & $\checkmark$ &  \textbf{68.1}\%	& \textbf{41.9}\%&	\textbf{64.7}\%\\
				\bottomrule[1.0pt]
			\end{tabular}
\vspace{-0.2in}
	\end{center}
\end{table}

\begin{table}[t]
	\begin{center}

		\caption{Aster~\cite{shi2018aster} accuracy of CCTBs with different Transformer Layout Strategies.}

		\label{table:CCTB_ablation_3}
			\begin{tabular}{c|cccc}
				\toprule[1.0pt]
			    Transformer Layout & easy& medium & hard & all \\
				\midrule[0.5pt]
				parallel+concat+1*1 conv & 75.4\% & 57.0\% & 40.0\%  & 58.6\%\\
				\midrule[0.5pt]
                horizontal-vertical &  76.2\% &	60.0\% & \textbf{41.6\%} & 60.4\% \\
                \midrule[0.5pt]
                vertical-horizontal & \textbf{76.8}\%	& \textbf{61.0}\% & 41.3\%	& \textbf{60.8}\%\\
				\bottomrule[1.0pt]
			\end{tabular}

	\end{center}
\end{table}

\begin{table*}[ht]
	\begin{center}
		\caption{Text recognition results of CRNN~\cite{crnn} model on six benchmark datasets. "Radius" denotes the radius of Gaussian kernels. "Preprocess" denotes the super-resolution method used before recognition('-' means BICUBIC interpolation).}
		\label{table:sync}
			\begin{tabular}{cc|ccccccc}
				\toprule[1.0pt]
				Radius & Preprocess & IC03~\cite{ic03} & IC13~\cite{ic13} & IIIK50~\cite{IIIK50} & SVT~\cite{svt} & CUTE80~\cite{cute80} & COCO~\cite{coco-text}\\
				\midrule[1.0pt]
				\multirow{4}*{0.5*} & - & 89.0\% & 82.5\% & 78.0\% & 73.1\% & 54.5\% & 36.2\%\\
				~ & +TSRN~\cite{wang2020scene} & 85.9\% & 80.5\% & 76.0\% & 70.3\% & 47.2\% & 35.2\%\\
				~ & +STT~\cite{scenetexttelescope}  & 87.2\% & 81.7\% & 77.3\% & 73.0\% & 51.0\% & 36.9\%\\
				~ & +TATT~\cite{TATT}  & 87.9\% & 82.6\% & 77.9\% & \textbf{75.4\%} & 50.4\% & 38.3\%\\
				~ & +TATSR & \textbf{89.3\%} & \textbf{83.4\%} & \textbf{78.2\%} & 74.7\% & \textbf{55.6\%} & \textbf{39.5\%}\\
				\midrule[1.0pt]
				\multirow{4}*{1} & - &  78.1\% & 68.9\% & 62.4\% & 52.9\% & 36.8\% & 24.8\%\\
				~ & +TSRN~\cite{wang2020scene} & 78.1\% & 70.1\% & 65.2\% & 51.0\% & 37.9\% & 26.9\%\\
				~ & +STT~\cite{scenetexttelescope} & 78.9\% & 72.9\% & 67.1\% & 57.5\% & 37.5\% & 29.2\%\\
				~ & +TATT~\cite{TATT}  & 80.2\% & \textbf{76.1\%} & 68.3\% & 61.2\% & 40.3\% & 31.2\%\\
				~ & +TATSR & \textbf{81.6\%} & 75.5\% & \textbf{69.9\%} & \textbf{62.6\%} & \textbf{41.7\%} & \textbf{32.6\%}\\
				\midrule[1.0pt]
				\multirow{4}*{1.5*} & - & 31.5\% & 24.6\% & 17.4\% & 11.8\% & 9.0\% & 5.3\%\\
				~ & +TSRN~\cite{wang2020scene} & 46.3\% & 40.2\% & 31.5\% & 20.1\% & 18.4\% & 8.8\%\\
				~ & +STT~\cite{scenetexttelescope} & 50.4\% & 44.0\%& 36.1\% & 26.4\% & 22.2\% & 11.4\%\\
				~ & +TATT~\cite{TATT}  & \textbf{61.8\%} & 52.6\% & 42.1\% & 33.5\% & 21.2\% & 14.2\%\\
				~ & +TATSR & 61.1\% & \textbf{55.0\%} & \textbf{45.1\%} & \textbf{34.8\%} & \textbf{22.2\%} & \textbf{15.2\%}\\
				\midrule[1.0pt]
				\multirow{4}*{2} & - & 4.0\% & 3.9\% & 3.3\% & 2.3\% & 1.0\% & 0.9\%\\
				~ & +TSRN~\cite{wang2020scene} & 14.9\% & 13.3\% & 10.3\% & 4.6\% & 7.6\% & 2.1\%\\
				~ & +STT~\cite{scenetexttelescope} & 20.3\% & 17.3\% & 13.6\% & 7.4\% & 6.9\% & 3.3\%\\
				~ & +TATT~\cite{TATT}  & 30.3\% & \textbf{27.5\%} & 19.8\% & 12.8\% & 8.3\% & 5.2\%\\
				~ & +TATSR & \textbf{31.7\%} & 27.0\% & \textbf{24.0\%} & \textbf{14.2\%} & \textbf{10.4\%} & \textbf{5.6\%}\\
				\bottomrule[1.0pt]
			\end{tabular}
	\end{center}
 \vspace{-0.4cm}
\end{table*}

\subsubsection{Interpretability Analysis of CCTB} To further understand our CCTB deeply, we replace the two transformers in CCTB with BLSTM and evaluate their recognition accuracy of short ($\mathrm{Len} < 9$) and long text samples ($\mathrm{Len} \geq 9$) in Tab.~\ref{table:CCTB_ablation_2}, respectively. Because of the stronger long-range modeling ability of transformer's multi-head attention than BLSTM's hidden state structure, compared to totally BLSTM-based, using only horizontal transformer achieves obvious improvement on long text samples, while the improvement on short texts is relatively small. 
Compared to horizontal transformer only, the vertical transformer in full CCTB enhances the recovery of the character shape, leading to a performance boost in the overall data. Such results show that our criss-cross strategy indeed separate the learning of different text priors into different directions and the sparse transformers indeed benefit the learning of each text prior.
We also compare the performance of different transformer layout strategies in Tab.~\ref{table:CCTB_ablation_3}. Because of more frequent alternative feature refinement, the vertical-first and horizontal-first strategies both outperform the parallel strategy. Since the order of the vertical and horizontal transformers has little effect on the results, we choose the one with the best results as the final setting.

\begin{figure}[t]  
	\includegraphics[width=0.49\textwidth]{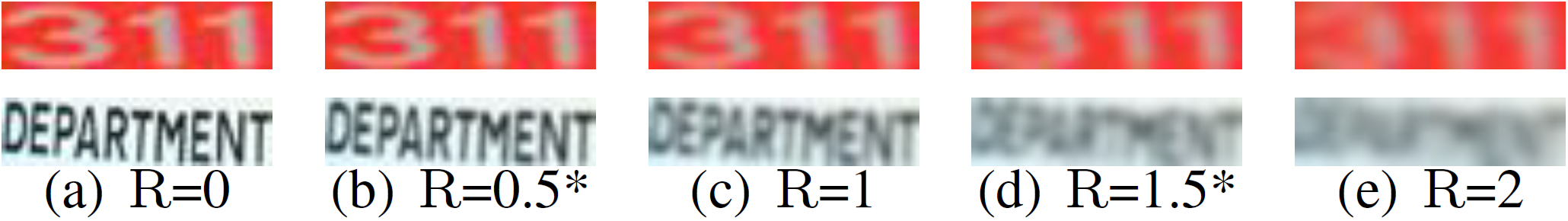}
	\caption{Example of images degraded by Gaussian blur kernels with different radii ($\mathrm{R}$). ($\mathrm{R}=1.5^{*}$ and $\mathrm{R}=0.5^{*}$ respectively means that we resize the images to $128\times 32$ first, then blur them with a Gaussian kernel with a radius of 3 or 1, and then resize them to $64\times16$)} 
	\label{fig:blur_example}
\end{figure}

\begin{figure*}[t] 
	\includegraphics[width=1\textwidth]{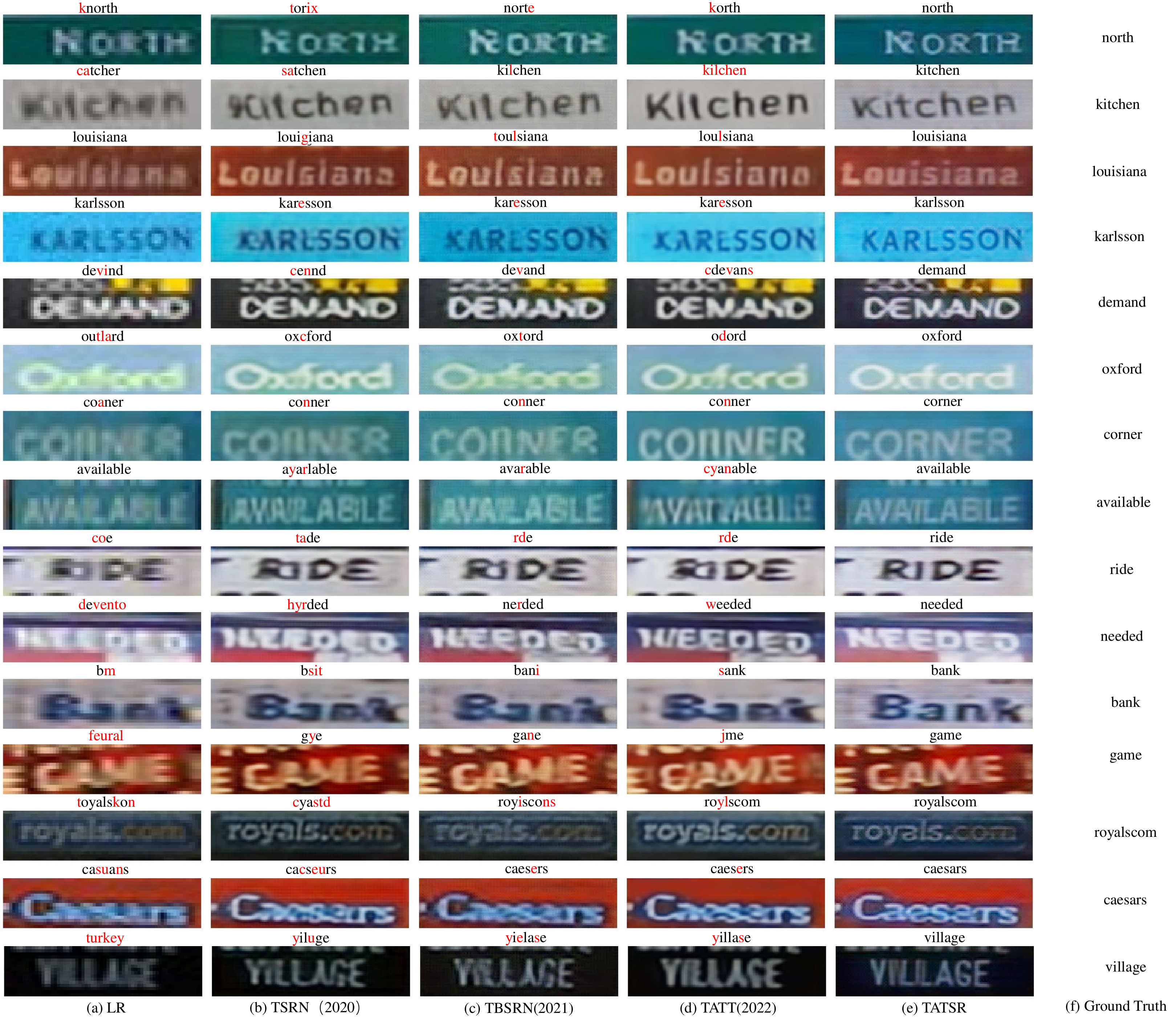}
    \vspace{-0.5cm}
	\caption{Examples of super-resolution results on real low-resolution images (no high-resolution ground truth). The first column shows the low-resolution images, and the four columns on the right show the super-resolution(SR) results. We offer the SR images with their text recognition results on the top. The last column shows the ground-truth texts. The text recognition results are predicted by CRNN~\cite{crnn}.} 
    
	\label{fig:real_low_cmp}
  \vspace{-0.2cm}
\end{figure*}

\subsection{Generalizability on Scene Text Recognition (STR) Benchmarks}
We conduct experiments on six STR benchmark datasets, including IC03~\cite{ic03}, IC13~\cite{ic13}, SVT~\cite{svt}, IIIK50~\cite{IIIK50}, CUTE80~\cite{cute80} and COCO~\cite{coco-text}, among which CUTE80~\cite{cute80} and COCO~\cite{coco-text} contain many samples with complex degradations, challenging for scene text image super-resolution.

\subsubsection{Datasets}
\noindent\textbf{IIIK50}~\cite{IIIK50} consists of 3000 test instances, taken from street scenes and originally-digital images.

\noindent\textbf{SVT}~\cite{svt}
SVT consists of 647 test images instances. Some images are severely degraded by noise, blur, and low resolution.

\noindent\textbf{IC03}~\cite{ic03}, \noindent\textbf{IC13}~\cite{ic13}
The datasets used in the Incidental Scene Text Competitions include 867 and 1015 test image instances, respectively.

\noindent\textbf{CUTE80}~\cite{cute80} consists of 288 test image instances. Since it focuses on curved text recognition, most images in CUTE80~\cite{cute80} have complex backgrounds, perspective distortion, and poor resolution.

\noindent\textbf{COCO}~\cite{coco-text} consist of 39K image instances(9835 for evaluation) cropped from the MS COCO~\cite{mscoco} dataset. Since MS COCO~\cite{mscoco} is not designed for text capturing, many cropped images in COCO~\cite{coco-text} contain occluded or low-resolution texts.

\subsubsection{Experimental Conditions}
For fairness, we use all methods trained on TextZoom~\cite{wang2020scene} dataset for comparison. Since STR datasets contain many high-resolution images in addition to low-resolution images, we conduct experiments under two different conditions: \textit{1)} All images with synthetic downsampling and degradation. \textit{2)} Real low-resolution images only.

In detail, for the first condition, we use all the images in six datasets. We use BICUBIC interpolation to resize images to $16 \times 64$. Then, we employ Gaussian blur kernels with different radii to imitate the degradation in real-world scenes(e.g., Fig.~\ref{fig:blur_example}). For the second condition, we choose images smaller than $16\times 64$ from all datasets and resize them to $16 \times 64$, a total of 3404 samples.

Under these two experimental conditions, we perform scene text recognition on the super-resolution images restored by different preprocessing methods, including BICUBIC interpolation, TSRN~\cite{wang2020scene}, TBSRN~\cite{scenetexttelescope}, TATT~\cite{TATT}, and our proposed TATSR.

\begin{table}[t]
    \vspace{-0.4cm}
	\begin{center}
		\caption{Recognition accuracy on real low-resolution images. "Preprocess" denotes the super-resolution method used before recognition('-' means BICUBIC interpolation).}
		\label{table:real}
			\begin{tabular}{c|ccc}
				\toprule[1.0pt]
				Preprocess & Aster~\cite{shi2018aster} & Moran~\cite{luo2019moran} & CRNN~\cite{crnn}\\
				\midrule[1.0pt]
				- & 58.3\% & 48.5\% & 41.1\%\\
				+TSRN~\cite{wang2020scene} & 58.5\% & 51.4\% & 40.0\%\\
				+STT~\cite{scenetexttelescope}  &  61.8\% & 54.2\% & 42.1\%\\
				+TATT~\cite{TATT}  &  62.1\% & 55.4\% & 43.5\%\\
				+TATSR & \textbf{62.8\%} & \textbf{55.8\%} & \textbf{44.8\%} \\
				\bottomrule[1.0pt]
			\end{tabular}
	\end{center}
 \vspace{-0.2in}
\end{table}

As shown in Tab.~\ref{table:sync}, all of the tested scene text image super-resolution methods visibly improve text recognition accuracy, especially on the heavily degraded images generated by large Gaussian kernels (e.g., $\mathrm{R}$=1.5* or 2). This proves the effectiveness of scene text image super-resolution as a preprocessing operation of STR. Among all of them, TATSR brings the biggest performance boost on all six benchmarks, showing the superiority of our method. In addition, it is worth noting that under the condition of small Gaussian kernels (\textit{e.g.,} $\mathrm{R}$=0.5* or 1), the text recognition accuracy of TSRN~\cite{wang2020scene} and TBSRN~\cite{scenetexttelescope} is even slightly inferior to the results of BICUBIC interpolation. We think this is caused by the domain gap between the TextZoom~\cite{wang2020scene} dataset and the STR datasets. In contrast, TATSR successfully enhances the images under all experimental conditions, showing better generalization.

Fig.~\ref{fig:real_low_cmp} and Tab.~\ref{table:real}  show the performance of different methods on real low-resolution scene text images.  After the pre-processing of TATSR, the accuracy increases by 4.53\% for Aster~\cite{shi2018aster}, 7.23\% for Moran~\cite{luo2019moran}, and 3.67\% for CRNN~\cite{crnn}. Compared with TSRN~\cite{wang2020scene}, TBSRN~\cite{scenetexttelescope}, and TATT~\cite{TATT}, the letters generated by TATSR in Fig.~\ref{fig:real_low_cmp} are clearer and more readable. Such results effectively verify our TATSR's superiority in enhancing the quality of low-resolution text images in real natural scenes, from both recognition accuracy and human perception.

\section{Conclusion}
This paper presented a new scene text image super-resolution (STISR) framework called TATSR. By calculating the similarity between recognition-based multi-scale features of high-resolution and super-resolution images, CP Loss overcomes the problems of previous text-oriented losses, which are the weak supervision of local structures, convergence conflict with pixel-wise losses, and cross-language unavailability. Meanwhile, a new sequence processing block called Criss-Cross Transformer Block (CCTB) is introduced to address the weakness of previous models on long text samples. Comprehensive experiments and ablation studies show that our TATSR framework can effectively improve the quality of low-resolution scene text images in term of both text recognition accuracy and human perception achieving a new state-of-the-art performance in STISR.

\nocite{*}
\bibliographystyle{IEEEtran}
\bibliography{egbib}

\newpage

\end{document}